\documentclass{article} 
\usepackage{iclr2019_conference,times}

\usepackage{booktabs} 
\usepackage{hyperref}
\usepackage{graphicx}
\usepackage{subfig}
\usepackage{csquotes} 
\usepackage{array, makecell}
\usepackage{amsfonts}
\usepackage{amsmath}
\usepackage{listings}
\usepackage{multirow}
\usepackage{color}
\usepackage{xcolor}
\usepackage{subfig}
\usepackage{diagbox}
\usepackage{appendix}
\usepackage{multirow}
\usepackage{algorithm2e}

\DeclareMathOperator{\wtm}{enc}
\newcommand{\embed}[1]{\boldsymbol{E}[#1]}

\newcommand{\absbest}[1]{\textbf{#1}}

\newcolumntype{Q}{>{\global\let\currentrowstyle\relax}}
\newcolumntype{^}{>{\currentrowstyle}}

\usepackage{amsmath,amsfonts,bm}

\def\eqref#1{equation~\ref{#1}}

\def\1{\bm{1}}

\def\mI{{\bm{I}}}

\DeclareMathAlphabet{\mathsfit}{\encodingdefault}{\sfdefault}{m}{sl}
\SetMathAlphabet{\mathsfit}{bold}{\encodingdefault}{\sfdefault}{bx}{n}

\title{CBOW Is Not All You Need: Combining CBOW with the Compositional
Matrix Space Model}

\author{Florian Mai \\
Idiap Research Institute\\
Martigny, Switzerland \\
\texttt{florian.mai@idiap.ch} \\
\And
Lukas Galke \\
Kiel University / ZBW\\
Germany \\
\texttt{lga@informatik.uni-kiel.de} \\
\And
Ansgar Scherp \\
University of Essex \\
United Kingdom \\
\texttt{ansgar.scherp@essex.ac.uk} 
}

\iclrfinalcopy{} 
\begin{document}

\maketitle

\begin{abstract}
Continuous Bag of Words (CBOW) is a powerful text embedding method. Due to
its strong capabilities to encode word content, CBOW embeddings perform well on
a wide range of downstream tasks while being efficient to compute. However,
CBOW is not capable of capturing the word order. The reason is that the
computation of CBOW's word embeddings is commutative, i.e., embeddings of XYZ
and ZYX are the same. In order to address this shortcoming, we propose a
learning algorithm for the Continuous Matrix Space
Model~\citep{rudolph2010compositional}, which we call Continual Multiplication
of Words (CMOW).
Our algorithm is an adaptation of
word2vec~\citep{mikolov2013efficient}, so that it can be trained on large
quantities of unlabeled text.
We empirically show that CMOW better captures linguistic properties, but it is
inferior to CBOW in memorizing word content. Motivated by these findings, we propose a
hybrid model that combines the strengths of CBOW and CMOW. Our results show
that the hybrid CBOW-CMOW-model retains CBOW's strong ability to memorize word
content while at the same time substantially improving its ability to encode
other linguistic information by 8\%. As a result, the hybrid also performs
better on 8 out of 11 supervised downstream tasks with an average improvement
of 1.2\%.
\end{abstract}

\section{Introduction}

Word embeddings are perceived as one of the most impactful contributions from
unsupervised representation learning to natural language processing from the
past few years~\citep{goth2016deep}.
Word embeddings are learned once on a large-scale stream of words.
A key benefit is that these pre-computed vectors can be re-used almost
universally in many different downstream applications. Recently, there has been
increasing interest in learning universal \emph{sentence} embeddings.
\cite{perone2018evaluation} have shown that the best encoding architectures are based on recurrent neural networks
(RNNs)~\citep{conneau2017supervised, peters2018deep} or the Transformer
architecture~\citep{cer2018universal}.
These techniques are, however, substantially more expensive to train and apply
than word embeddings~\citep{dblp:conf/naacl/hillck16,cer2018universal}. 
Their usefulness is therefore limited when fast processing of large volumes of data is critical. 

More efficient encoding techniques are typically based on aggregated word
embeddings such as Continuous Bag of Words (CBOW), which is a mere summation of the word
vectors~\citep{mikolov2013efficient}.
Despite CBOW's simplicity, it attains strong results on many downstream
tasks. Using sophisticated weighting schemes, the performance of aggregated word
embeddings can be further increased~\citep{arora2016simple}, coming even close to strong
LSTM baselines~\citep{ruckle2018concatenated, DBLP:conf/acl/HenaoLCSSWWMZ18}
such as InferSent~\citep{conneau2017supervised}. This raises the question how
much benefit recurrent encoders actually provide over simple word embedding
based methods~\citep{anonymous2018no}. In their analysis,
\citet{DBLP:conf/acl/HenaoLCSSWWMZ18} suggest that the main difference may be
the ability to encode word order.
In this paper, we propose an intuitive method to enhance aggregated word
embeddings by word order awareness.

The major drawback of these CBOW-like approaches is that they are
solely based on addition. However, \emph{addition is not all you need}.
Since it is a commutative operation, the aforementioned methods are not able to 
capture any notion of word order. 
However, word order information is crucial for some tasks, e.g., sentiment
analysis~\citep{DBLP:conf/acl/HenaoLCSSWWMZ18}.
For instance, the following two sentences yield the exact same embedding in an
addition-based word embedding aggregation technique: ``The movie was not awful,
it was rather great.'' and ``The movie was not great, it was rather awful.''
A classifier based on the CBOW embedding of these sentences would inevitably
fail to distinguish the two different
meanings~\cite[p. 151]{DBLP:series/synthesis/2017Goldberg}.

To alleviate this drawback, \citet{rudolph2010compositional} propose to model
each word as a matrix rather than a vector, and compose multiple word embeddings
\emph{via matrix multiplication rather than addition}. This so-called
\emph{Compositional Matrix Space Model} (CMSM) of language has powerful
theoretical properties that subsume properties from vector-based models and
symbolic approaches. The most obvious advantage is the non-commutativity of
matrix multiplication as opposed to addition, which results in order-aware encodings.

In contrast to vector-based word embeddings, there is so far no
solution to effectively train the parameters of word matrices on large-scale
\emph{unlabeled} data.
Training schemes from previous work were specifically designed for
sentiment analysis~\citep{yessenalina2011compositional, asaadi2017gradual}.
Those require complex, multi-stage initialization, which indicates the
difficulty of training CMSMs.
We show that CMSMs can be trained in a similar way as the well-known
CBOW model of word2vec~\citep{mikolov2013efficient}. We make two simple yet critical
changes to the initialization strategy and training objective of CBOW\@. Hence, we present the
first unsupervised training scheme for CMSMs, which we call 
\emph{Continual Multiplication Of Words} (CMOW).

We evaluate our model's capability to capture linguistic properties in the
encoded text. We find that CMOW and CBOW have properties that are
complementary. On the one hand, CBOW yields much stronger results at the word
content memorization task. CMOW,
on the other hand, offers an advantage in all other linguistic probing tasks, often by a wide margin. Thus,
we propose a hybrid model to jointly learn the word vectors of CBOW and the
word matrices for CMOW\@.

Our experimental results confirm the effectiveness of our hybrid CBOW-CMOW
approach. At comparable embedding size, CBOW-CMOW retains CBOW's
ability to memorize word content while at the same time improves the
performance on the linguistic probing tasks by 8\%.
CBOW-CMOW outperforms CBOW at 8 out of 11 supervised downstream tasks scoring
only 0.6\% lower on the tasks where CBOW is slightly better. On average, the
hybrid model improves the performance over CBOW by 1.2\% on supervised
downstream tasks, and by 0.5\% on the unsupervised tasks.

In summary, our contributions are:
(1) For the first time, we present an unsupervised, efficient training
  scheme for the Compositional Matrix Space Model. Key elements of our scheme are
  an initialization strategy and training objective that are specifically
  designed for training CMSMs.
  (2) We quantitatively demonstrate that the strengths of the resulting
  embedding model are complementary to classical CBOW embeddings.
  (3) We successfully combine both approaches into a hybrid model that is
  superior to its individual parts.

After giving a brief overview of the related work, we formally introduce CBOW,
CMOW, and the hybrid model in Section~\ref{sec:encoder}.
We describe our experimental setup and present the results in
Section~\ref{sec:experiments}. The results are discussed in
Section~\ref{sec:discussion}, before we conclude.

\section{Related work}\label{sec:related-work}
We present an algorithm for learning the weights of the
Compositional Matrix Space Model~\citep{rudolph2010compositional}.
To the best of our knowledge, only \cite{yessenalina2011compositional} and
\cite{asaadi2017gradual} have addressed this. They present complex,
multi-level initialization strategies to achieve reasonable results. Both papers
train and evaluate their model on sentiment analysis datasets only, but they do
not evaluate their CMSM as a general-purpose sentence encoder.

Other works have represented words as matrices as well, but unlike our work not
within the framework of the CMSM. \cite{grefenstette2011experimental} represent
only relational words as matrices. \cite{socher2012semantic} and
\cite{chung2017lifted} argue that while CMSMs are arguably more expressive than
embeddings located in a vector space, the associativeness of matrix
multiplication does not reflect the hierarchical structure of language. Instead,
they represent the word sequence as a tree structure. \cite{socher2012semantic}
directly represent each word as a matrix (and a vector) in a recursive neural
network.  \cite{chung2017lifted} present a two-layer architecture. In the first
layer, pre-trained word embeddings are mapped to their matrix representation. In
the second layer, a non-linear function composes the constituents.

Sentence embeddings have recently become an active field of
research. A desirable property of the embeddings is that the encoded
knowledge is useful in a variety of high-level downstream
tasks. To this end, \cite{conneau2018senteval} and \cite{conneau2018you}
introduced an evaluation framework for sentence encoders that tests both their
performance on downstream tasks as well as their ability to capture
linguistic properties. Most works focus on either i) the \emph{ability} of
encoders to capture appropriate semantics or on ii) training objectives that
give the encoders \emph{incentive} to capture those semantics. Regarding the
former, large RNNs are by far the most popular~\citep{conneau2017supervised,
kiros2015skip, tang2017rethinking, nie2017dissent, dblp:conf/naacl/hillck16, mccann2017learned,
peters2018deep, logeswaran2018an}, followed by convolutional neural
networks~\citep{gan2017learning}.
A third group are efficient methods that aggregate word
embeddings~\citep{wieting2015towards, arora2016simple,
pagliardini2017unsupervised, ruckle2018concatenated}. Most of the methods in the
latter group are word order agnostic.
Sent2Vec~\citep{pagliardini2017unsupervised} is an exception in the sense that
they also incorporate bigrams. Despite also employing an objective similar to
CBOW, their work is very different to ours in that they still use addition as
composition function.
Regarding the training objectives, there is an ongoing debate whether language
modeling~\citep{peters2018deep, DBLP:conf/acl/RuderH18}, machine
translation~\citep{mccann2017learned}, natural language
inference~\citep{conneau2017supervised}, paraphrase
identification~\citep{wieting2015towards}, or a mix of many
tasks~\citep{subramanian2018learning} is most appropriate for incentivizing the
models to learn important aspects of language. In our study, we
focus on adapting the well-known objective from
word2vec~\citep{mikolov2013efficient} for the CMSM.

\section{Methods: CBOW and CMOW}\label{sec:encoder}

We formally present CBOW and CMOW encoders in a unified framework.
Subsequently, we discuss the training objective, the initialization strategy,
and the hybrid model.

\subsection{Text encoding}

We start with a lookup table for the word matrices, i.e., an embedding,
$\boldsymbol{E} \in \mathbb{R}^{m \times d \times d}$, where $m$ is the
vocabulary size and $d$ is the dimensionality of the (square) matrices.
We denote a specific word matrix of the embedding by $\embed{\cdot}$.
By $\Delta \in \{\sum, \prod\}$ we denote the function that aggregates word
embeddings into a sentence embedding. Formally, given a sequence $s$ of
arbitrary length $n$, the sequence is encoded as $\Delta_{i=1}^n \embed{s_i}$.
For $\Delta = \sum$, the model becomes CBOW. By setting $\Delta =
\prod$ (matrix multiplication), we obtain CMOW. Because the result of
the aggregation for any prefix of the sequence is again a square
matrix of shape $d \times d$ irrespective of the aggregation function, the
model is well defined for any non-zero sequence length. Thus, it can serve as
a general-purpose text encoder.

Throughout the remainder of this paper, we denote the encoding step by
$\wtm^{\boldsymbol{E}}_{\Delta} (s) := \operatorname{flatten}\left(
\Delta_{i=1}^n \embed{s_i} \right)$, where
$\operatorname{flatten}$ concatenates the columns of the matrices to obtain a
vector that can be passed to the next layer.

\subsection{Training objective}\label{sub:cbow}
Motivated by its success, we employ a similar training objective as
word2vec~\citep{dblp:conf/nips/mikolovsccd13}.
The objective consists of maximizing the conditional
probability of a word $w_O$ in a certain context $s$: $p(w_O \mid s)$.
For a word $w_t$ at position $t$ within a sentence, we consider the window of
tokens $\left(w_{t-c}, \dotsc, w_{t+c} \right)$ around that word. From that
window, a target word $w_O := \left\{ w_{t + i} \right\}, i \in \{-c, \ldots, +
c\}$ is selected. The remaining $2c$ words in the window are used as the context $s$.
The training itself is conducted via negative sampling $\operatorname{NEG-k}$,
which is an efficient approximation of the softmax~\citep{dblp:conf/nips/mikolovsccd13}.
For each positive example, $k$ negative examples (noise words) are drawn from
some noise distribution $P_n(w)$.
The goal is to distinguish the target word $w_O$ from the
randomly sampled noise words. Given the encoded input words $\wtm^{}_{\Delta}
(s)$, a logistic regression with weights $v \in \mathbb{R}^{m \times d^2}$ is
conducted to predict $1$ for context words and $0$ for noise words. The negative
sampling training objective becomes:

\begin{align}\label{eq:negative-sampling-single}
  \log \sigma \left( v_{w_O}^T \wtm^{\boldsymbol{E}}_{\Delta} (s) \right)  +
  \sum_{i=1}^{k} \mathbb{E}_{w_i \sim {P_n}(w)} \left\lbrack \log \sigma \left( - v_{w_i}^T
    \wtm^{\boldsymbol{E}}_{\Delta} (s) \right) \right\rbrack
\end{align}

In the original
word2vec~\citep{mikolov2013efficient}, the center word $w_O := w_{t}$ is
used as the target word. In our experiments, however, this objective did not
yield to satisfactory results. We hypothesize that this objective is too easy to
solve for a word order-aware text encoder, which diminishes incentive for the
encoder to capture semantic information at the sentence level.
Instead, we propose to select a random output word $w_O \sim \mathcal{U} (\{w_{t-c}, \dotsc,
w_{t+c} \})$ from the window. The rationale is the following: By removing
the information at which position the word was removed from the window, the
model is forced to build a semantically rich representation of the \emph{whole}
sentence. For CMOW, modifying the objective leads to a large improvement on
downstream tasks by 20.8\% on average, while it does not make a difference for
CBOW. We present details in the appendix
(Section~\ref{app:compare-objectives}).

\subsection{Initialization}\label{sec:initialization}

So far, only \cite{yessenalina2011compositional} and \cite{
asaadi2017gradual} have proposed algorithms for learning the parameters for the matrices in
CMSMs. Both works devote
particular attention to the initialization, noting that a standard
initialization randomly sampled from $\mathcal{N} ( 0 , 0.1 )$ does not work
well due to the optimization problem being non-convex. To alleviate this, the
authors of both papers propose rather complicated initialization strategies
based on a bag-of-words solution~\citep{yessenalina2011compositional} or incremental
training, starting with two word phrases~\citep{asaadi2017gradual}.
We instead propose an effective yet simple strategy, in which the
embedding matrices are initialized close to the identity matrix.

\begin{figure*}
\center

\includegraphics[width=0.7\textwidth]{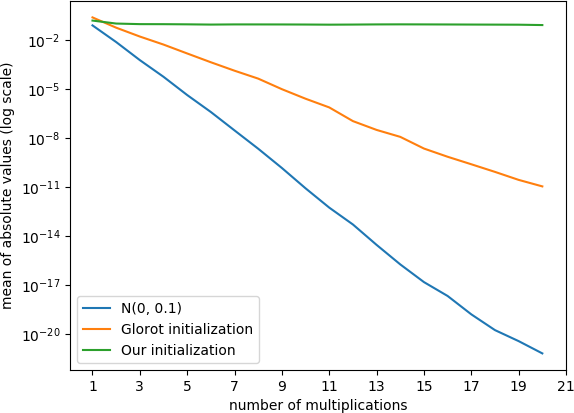}
\caption{Mean of the absolute values of the text embeddings (y-axis)
plotted depending on the number of multiplications (x-axis) for the three
initialization strategies. As one can see, the absolute value of the embeddings
sharply decreases for the initialization strategies Glorot and
$\mathcal{N}(0, 0.1)$ the more multiplications are performed. In contrast, when our initialization method
is applied, the absolute values of the embeddings have the same magnitude
regardless of the sentence length.
}
\label{fig:vanishing_values}
\end{figure*}

We argue that modern optimizers based on stochastic gradient descent have proven
to find good solutions to optimization problems even when those are non-convex
as in optimizing the weights of deep neural networks.
CMOW is essentially a deep linear neural
network with flexible layers, where each layer corresponds to a word in the
sentence. The output of the final layer is then used as an embedding for the
sentence. A subsequent classifier may expect that all embeddings come from the same distribution.
We argue that initializing the weights
randomly from $\mathcal{N} ( 0 , 0.1 )$ or any other distribution that has most
of its mass around zero is problematic in such a setting. This includes the
Glorot initialization~\citep{glorot2010understanding}, which was designed to
alleviate the problem of vanishing gradients. Figure~\ref{fig:vanishing_values}
illustrates the problem: With each multiplication, the values in the embedding
become smaller (by about one order of magnitude).
This leads to the undesirable effect that short sentences have a drastically
different representation than larger ones, and that the embedding values vanish
for long sequences.

To prevent this problem of vanishing values, we propose an initialization strategy, where each
word embedding matrix $\embed{w} \in \mathbb{R}^{d \times d}$ is initialized as
a random deviation from the identity matrix:

\begin{align*}
  \embed{w} := 
\begin{pmatrix}
 \mathcal{N} ( 0, 0.1 ) & \ldots &  \mathcal{N} ( 0
, 0.1 ) \\
\vdots & \ddots & \vdots \\
\mathcal{N} ( 0, 0.1 ) & \ldots & \mathcal{N} ( 0
, 0.1 )
\end{pmatrix}
 + \mI_{d},
\end{align*}

It is intuitive and also easy to prove that the expected value of the
multiplication of any number of such word embedding matrices is again the
identity matrix (see Appendix~\ref{appendix:proof}).
Figure~\ref{fig:vanishing_values} shows how our initialization strategy
is able to prevent vanishing values.
For training CMSMs, we observe a substantial improvement over Glorot
initialization of 2.8\% on average. We present details in
Section~\ref{app:sec:init} of the appendix.

\subsection{Hybrid CBOW-CMOW model}

Due to their different nature, CBOW and CMOW also capture different
linguistic features from the text. It is therefore intuitive to expect that a
hybrid model that combines the features of their constituent models
also improves the performance on downstream tasks. 

The simplest combination is to train CBOW and CMOW
separately and concatenate the resulting sentence embeddings at test time.
However, we did not find this approach to work well in preliminary
experiments.
We conjecture that there is still a considerable overlap in the features learned by each model, which
hinders better performance on downstream tasks.
To prevent redundancy in the learned features, we expose CBOW and CMOW to a
shared learning signal by training them jointly. To this end, we modify
Equation~\ref{eq:negative-sampling-single} as follows:

\begin{align*}\label{eq:negative-sampling-combined}
  \log \sigma \left( v_{w_O}^T \lbrack\wtm^{\boldsymbol{E_1}}_{\sum} (s)
  ; \wtm^{\boldsymbol{E_2}}_{\prod} (s)\rbrack \right) \\
  + \sum_{i=1}^{k} \mathbb{E}_{w_i \sim {P_n}(w)} \left\lbrack \log \sigma
  \left( - v_{w_i}^T \lbrack\wtm^{\boldsymbol{E_1}}_{\sum} (s)
  ; \wtm^{\boldsymbol{E_2}}_{\prod} (s)\rbrack \right) \right\rbrack.
\end{align*}
Intuitively, the model uses logistic regression to predict the missing word
from the concatenation of CBOW and CMOW embeddings. Again,
$\boldsymbol{E_i} \in \mathbb{R}^{m \times d_i \times d_i}$ are separate word
lookup tables for CBOW and CMOW, respectively, and $v \in \mathbb{R}^{m \times
(d_{1}^2 + d_{2}^2)}$ are the weights of the logistic regression.

\section{Experiments}\label{sec:experiments}

We conducted experiments to evaluate the effect of using our proposed models for
training CMSMs. In this section, we describe the experimental setup and present
the results on linguistic probing as well as downstream tasks.

\subsection{Experimental setup}
In order to limit the total batch size and to avoid expensive
tokenization steps as much as possible, we created each batch in the following
way:
1,024 sentences from the corpus are selected at random. After tokenizing each
sentence, we randomly select (without replacement) at maximum 30 words from the
sentence to function as center words for a context window of size $c = 5$,
i.e., we generate up to 30 training samples per sentence. By padding with
copies of the neutral element, we also include words as center words for which
there are not enough words in the left or the right context. For CBOW, the
neutral element is the zero matrix. For CMOW, the neutral element is the identity matrix.

We trained our models on the unlabeled UMBC news
corpus~\citep{han2013umbc_ebiquity}, which consists of about 134 million
sentences and 3 billion tokens. Each sentence has 24.8 words on average with a
standard deviation of 14.6. Since we only draw 30 samples per sentence to
limit the batch size, not all possible training examples are used in an epoch,
which may result in slightly worse generalization if the model is trained for a
fixed number of epochs. We therefore use 0.1\% of the 134 million sentences for
validation. After 1,000 updates (i.e., approximately every millionth training
sample) the validation loss is calculated, and training terminates after 10
consecutive validations of no improvement.
Following \cite{dblp:conf/nips/mikolovsccd13}, we limit the vocabulary to the
30,000 most-frequent words for comparing our different methods and their variants.
Out-of-vocabulary words are discarded. The optimization is carried out by
Adam~\citep{adam} with an initial learning rate of 0.0003 and $k = 20$ negative
samples as suggested by \cite{dblp:conf/nips/mikolovsccd13} for rather small datasets.
For the noise distribution $P_n(w)$ we again follow
\cite{dblp:conf/nips/mikolovsccd13} and use $\mathcal{U}(w)^{3/4}/Z$, where $Z$
is the partition function to normalize the distribution.

We have trained five different models: CBOW and CMOW with $d =
20$ and $d = 28$, which lead to 400-dimensional and
784-dimensional word embeddings, respectively. We also trained
the Hybrid CBOW-CMOW model with $d = 20$ for each component, so that the total
model has 800 parameters per word in the lookup tables. We report
the results of two more models: H-CBOW is the 400-dimensional
CBOW component trained in Hybrid and H-CMOW is the
respective CMOW component. Below, we compare the 800-dimensional
Hybrid method to the 784-dimensional CBOW and CMOW
models.

After training, only the encoder of the model 
$\wtm^{\boldsymbol{E}}_{\Delta}$ is retained. We assess the
capability to encode linguistic properties by evaluating on 10 linguistic
probing tasks~\citep{conneau2018you}. In particular, the Word Content (WC) task
tests the ability to memorize exact words in the sentence. Bigram Shift (BShift)
analyzes the encoder's sensitivity to word order. The downstream performance is
evaluated on 10 supervised and 6 unsupervised tasks
from the SentEval framework~\citep{conneau2018senteval}. We use the standard
evaluation configuration, where a logistic regression classifier is trained on top of the embeddings.

\subsection{Results on linguistic probing tasks}

Considering the linguistic probing tasks (see
Table~\ref{tab:cmow-cbow-comp-probing}), CBOW and CMOW show complementary results.
While CBOW yields the highest performance at word content memorization,
CMOW outperforms CBOW at all other tasks.
Most improvements vary between 1-3 percentage points. The difference is
approximately 8 points for CoordInv and Length, and even 21 points for BShift.

The hybrid model yields scores close to or even above the better model of the
two on all tasks. In terms of relative numbers, the hybrid model improves upon CBOW in
all probing tasks but WC and SOMO. The relative improvement averaged
over all tasks is 8\%. Compared to CMOW, the hybrid model shows rather small differences.
The largest loss is by 4\% on the CoordInv task. However, due to the large gain
in WC (20.9\%), the overall average gain is still 1.6\%.

We now compare the jointly trained H-CMOW and H-CBOW with their separately
trained 400-dimensional counterparts. We observe that CMOW
loses most of its ability to memorize word content, while CBOW shows a slight
gain. On the other side, H-CMOW shows, among others, improvements
at BShift.

\begin{table*}
\caption{Scores on the probing tasks attained by our models.
Rows starting with ``Cmp.'' show the relative change with respect to Hybrid.}
\label{tab:cmow-cbow-comp-probing}
\resizebox{\textwidth}{!}{
\begin{tabular}{Ql|^l|^r^r^r^r^r^r^r^r^r^r}
\toprule
 Dim &  Method &  Depth &  BShift &  SubjNum &  Tense &  CoordInv & 
   Length &  ObjNum &  TopConst &  SOMO &  WC \\
\midrule
\multirow{4}{*}{400} & CBOW/400 &   32.5 &         50.2 &        78.9 &   78.7 &          
53.6 &    73.6 &       79.0 &             69.6 &       48.9 &         86.7 \\
& CMOW/400 &  \absbest{34.4} &         68.8 &        80.1 &   \absbest{79.9} &                  
\absbest{59.8} &    81.9 &       \absbest{79.2} &             \absbest{70.7} &   
\absbest{50.3} & 70.7 \\
 & H-CBOW &   31.2 &         50.2 &        77.2 &   78.8 &                  
52.6 & 77.5 &       76.1 &             66.1 &       49.2 &        
\absbest{87.2}
\\
& H-CMOW &   32.3 &         \absbest{70.8} &        \absbest{81.3} &   76.0 &                  
59.6 & \absbest{82.3} &       77.4 &             70.0 &       50.2 &        
38.2
\\
\midrule
\midrule
\multirow{ 2}{*}{784} & CBOW/784 &   33.0 &         49.6 &        79.3 &   78.4
& 53.6 & 74.5 &       78.6 &             72.0 &       49.6 &        
\absbest{89.5} \\
 & CMOW/784 &   \absbest{35.1} &         \absbest{70.8} &       
 \absbest{82.0} & 80.2 & \absbest{61.8} & 82.8 &       \absbest{79.7} &             74.2 &      
 \absbest{50.7} & 72.9 \\
 800 & Hybrid &   35.0 &         \absbest{70.8} &        81.7 &  
 \absbest{81.0} & 59.4 & \absbest{84.4} &       79.0 &            
 \absbest{74.3} &       49.3 & 87.6 \\
- & cmp. CBOW &   +6.1\% &      +42.7\%    &  +3\%  &  +3.3\% &                  
 +10.8\% & +13.3\% & +0.5\% & +3.2\% & -0.6\%  & -2.1\% \\
- & cmp. CMOW & -0.3\%  &  +-0\%   &  -0.4\%  & +1\% &                  
 -3.9\% & +1.9\% & -0.9\% & +0.1\% & -2.8\% & +20.9\% \\
\bottomrule
\end{tabular}
}
\end{table*}

\subsection{Results on downstream tasks}

Table~\ref{tab:cmow-cbow-comp-downstream-supervised} shows the scores from the
supervised downstream tasks. Comparing the 784-dimensional models, again, CBOW
and CMOW seem to complement each other. This time, however, CBOW has the
upperhand, matching or outperforming CMOW on all supervised downstream tasks
except TREC by up to 4 points. On the TREC task, on the other hand, CMOW
outperforms CBOW by 2.5 points.

Our jointly trained model is not more than
0.8 points below the better one of CBOW and CMOW on any of the considered
supervised downstream tasks. On 7 out of 11 supervised tasks,
the joint model even improves upon the better model, and on SST2, SST5, and MRPC
the difference is more than 1 point. The average relative improvement over all
tasks is 1.2\%.

Regarding the unsupervised downstream tasks
(Table~\ref{tab:cmow-cbow-comp-downstreamU}), CBOW is clearly superior to
CMOW on all datasets by wide margins. For example, on STS13, CBOW's score is
50\% higher. The hybrid model is able to repair this deficit, reducing the
difference to 8\%. It even outperforms CBOW on two of the tasks, and
yields a slight improvement of 0.5\% on average over all unsupervised
downstream tasks. However, the variance in relative performance is notably
larger than on the supervised downstream tasks.

\begin{table*}
\caption{Scores on supervised downstream tasks attained by our models.
Rows starting with ``Cmp.'' show the relative change with respect to Hybrid.
}
\label{tab:cmow-cbow-comp-downstream-supervised}
\resizebox{\textwidth}{!}{
\begin{tabular}{Ql^l^r^r^r^r^r^r^r^r^r^r^r}
\toprule
   Method & SUBJ &    CR &    MR &  MPQA &  MRPC &  TREC & 
   SICK-E &  SST2 &  SST5 &  STS-B &  SICK-R \\
\midrule
 CBOW/784 &   90.0 &  \absbest{79.2} &  \absbest{74.0} &  87.1 &  71.6 &  85.6 &           
 78.9 & 78.5 &  42.1 &          61.0 &             \absbest{78.1} \\
 CMOW/784 &   87.5 &  73.4 &  70.6 &  \absbest{87.3} &  69.6 &  \absbest{88.0} &           
 77.2 & 74.7 &  37.9 &          56.5 &             76.2 \\
 Hybrid &    \absbest{90.2} &  78.7 &  73.7 &  \absbest{87.3} &  \absbest{72.7}
 & 87.6 & \absbest{79.4} & \absbest{79.6} &  \absbest{43.3} &         
 \absbest{63.4} & 77.8
 \\
 cmp. CBOW &  +0.2\%   & -0.6\%  & -0.4\% & +0.2\%  &  +1.5\% & +2.3\%  &            
 +0.6\% & +1.4\% & +2.9\%  & +3.9\%   & -0.4\% \\
 cmp. CMOW & +3.1\%   & +7.2\%  &  +4.4\% &  +0\% & +4.5\%  & -0.5\% &            
 +2.9\% & +6.7\% & +14.3  &       +12.2\%    &    +2.1\%          \\
\bottomrule
\end{tabular}

}
\end{table*}

\begin{table*}
\caption{Scores on unsupervised downstream tasks attained by our models.
Rows starting with ``Cmp.'' show the relative change with respect to Hybrid.}
\label{tab:cmow-cbow-comp-downstreamU}
\center
\resizebox{0.6\textwidth}{!}{
\begin{tabular}{Ql^r^r^r^r^r}
\toprule
   Method &  STS12 &  STS13 &  STS14 &  STS15 &  STS16 \\
\midrule
 CBOW &   43.5 &   \absbest{50.0} &   \absbest{57.7} &   \absbest{63.2} &   61.0
 \\
 CMOW &   39.2 &   31.9 &   38.7 &   49.7 &   52.2 \\
 Hybrid &   \absbest{49.6} &   46.0 &   55.1 &   62.4 &   \absbest{62.1} \\
 cmp. CBOW &   +14.6\% & -8\% & -4.5\%  &  -1.5\% & +1.8\%  \\
 cmp. CMOW &   +26.5\% & +44.2\%  &  +42.4 &  +25.6\% & +19.0\% \\
\bottomrule
\end{tabular}}
\end{table*}

\section{Discussion}\label{sec:discussion}

Our CMOW model produces sentence embeddings that are approximately at the level
of fastSent~\citep{dblp:conf/naacl/hillck16}.
Thus, CMOW is a reasonable choice as a sentence encoder.
Essential to the success of our training schema for the CMOW model are two
changes to the original word2vec training.
First, our initialization strategy improved the downstream performance by
2.8\% compared to Glorot initialization.
Secondly, by choosing the target word of the objective at random, the
performance of CMOW on downstream tasks improved by 20.8\% on average.
Hence, our novel training scheme is the first that provides an effective way to
obtain parameters for the Compositional Matrix Space Model of language from unlabeled, large-scale datasets.

Regarding the probing tasks, we observe that CMOW embeddings
better encode the linguistic properties of sentences than CBOW.
CMOW gets reasonably close to CBOW on some downstream tasks.
However, CMOW does not in general supersede CBOW embeddings.
This can be explained by the fact that CBOW is stronger at word content
memorization, which is known to highly correlate with the performance on most
downstream tasks~\citep{conneau2018you}.
Yet, CMOW has an increased performance on the TREC question type
classification task ($88.0$ compared to $85.6$).
The rationale is that this particular TREC task belongs to a class of downstream
tasks that require capturing other linguistic properties apart from Word
Content~\citep{conneau2018you}.

Due to joint training, our hybrid model learns to pick up the best features from
CBOW and CMOW simultaneously. It enables both models to focus on their
respective strengths. This can best be seen by observing that H-CMOW almost
completely loses its ability to memorize word content. In return, H-CMOW has
more capacity to learn other properties, as seen in the increase in performance at
BShift and others. A complementary behavior can be observed for H-CBOW, whose
scores on Word Content are increased. Consequently, with an 8\% improvement on
average, the hybrid model is substantially more linguistically informed than
CBOW.
This transfers to an overall performance improvement by 1.2\% on
average over 11 supervised downstream tasks, with large
improvements on sentiment analysis tasks (SST2, SST5), question classification
(TREC), and the sentence representation benchmark (STS-B). The improvements on
these tasks is expected because they arguably depend on word order information.
On the other tasks, the differences are small. Again, this can be explained by
the fact that most tasks in the SentEval framework mainly depend on word content
memorization~\citep{conneau2018you}, where the hybrid model does not improve
upon CBOW.

Please note, the models in our study do not represent the state-of-the-art for
sentence embeddings. ~\citet{perone2018evaluation} show that better scores are
achieved by LSTMs and Transformer models, but also by averaging word embedding
from fastText~\citep{mikolov2017advances}.
These embeddings were trained on the CBOW objective, and are thus very similar
to our models. However, they are trained on large corpora (600B tokens vs 3B in
our study), use large vocabularies (2M vs 30k in our study), and incorporate
numerous tricks to further enhance the quality of their models: word subsampling,
subword-information, phrase representation, n-gram representations,
position-dependent weighting, and corpus de-duplication. In the present study,
we focus on comparing CBOW, CMOW, and the hybrid model in a scenario where we
have full control over the independent variables.
To single out the effect of the independent variables better, we keep our
models relatively simple.
Our analysis yields interesting insights on what our models learn when trained
separately or jointly, which we consider more valuable in the long term for the
research field of text representation learning.

We offer an efficient order-aware extension to embedding algorithms from the bag-of-words family. 
Our 784-dimensional CMOW embeddings can be computed at the same rate as CBOW embeddings.
We empirically measured in our experiments 71k for CMOW vs. 61k for CBOW in terms of encoding sentences per second.
This is because of the fast implementation of matrix multiplication in GPUs.
It allows us to encode sentences approximately 5 times faster than using a simple Elman RNN of the same size (12k per second).
Our matrix embedding approach also offers valuable theoretical advantages
over RNNs and other autoregressive models. Matrix multiplication is associative
such that only $\log_2 n$ sequential steps are necessary to encode a sequence
of size $n$. Besides parallelization, also dynamic programming techniques can
be employed to further reduce the number of matrix multiplication steps, e.~g.,
by pre-computing frequent bigrams. We therefore expect our matrix embedding
approach to be specifically well-suited for large-scale, time-sensitive text
encoding applications. 
Our hybrid model serves as a blueprint for using CMOW in conjunction
with other existing embedding techniques such as fastText~\citep{mikolov2017advances}.

\section{Conclusion}\label{sec:conclusion}
We have presented the first efficient, unsupervised learning scheme
for the word order aware Compositional Matrix Space Model.
We showed that the resulting sentence embeddings
capture linguistic features that are complementary to CBOW embeddings. We thereupon
presented a hybrid model with CBOW that is able to combine the complementary
strengths of both models to yield an improved downstream task performance, in
particular on tasks that depend on word order information. Thus, our model
narrows the gap in terms of representational power between simple word
embedding based sentence encoders and highly non-linear recurrent sentence encoders.

We made the code for this paper available at
\url{https://github.com/florianmai/word2mat}.

\subsubsection*{Acknowledgement}
This research was supported by the Swiss National Science Foundation under the project Learning
Representations of Abstraction for Opinion Summarisation (LAOS), grant number ``FNS-30216''.

\bibliography{main}
\bibliographystyle{iclr2019_conference}

\newpage

\begin{appendices}
\section*{Appendices}
\section{Proof of constant expected value of matrix
multiplication}\label{appendix:proof} The statement that we formally proof is
the following. For any sequence $s = s_1\ldots s_n$:
\begin{equation*}
\forall 1 \leq k \leq n: \mathbb{E}[\wtm_{\prod}(s_1,\ldots,s_k)] =
\mI_{d}.
\end{equation*}

The basis ($n = 1$) follows trivially due to the expected value of each entry
being the mean of the normal distribution. For the induction step, let
$\mathbb{E}[ \prod\limits_{i = 1}^{n} ( W_i )] = \mI_{d}$. It follows:
\begin{align*}
  & \mathbb{E}[ \prod\limits_{i = 1}^{n + 1} ( W_i
)] &\\
= & \mathbb{E}[ \prod\limits_{i = 1}^{n} ( W_i
) \cdot W_{n + 1}] & \\
= & \mathbb{E}[ \prod\limits_{i = 1}^{n} ( W_i
)] \cdot \mathbb{E}[W_{n + 1}] & \text{(Independence)} \\
= & \mI_{d} \cdot \mathbb{E}[W_{n + 1}] & \text{(Hypothesis)} \\
= & \mI_{d} \cdot \mI_{d} & \text{(Exp. val of each entry)} \\
= & \mI_{d}
\end{align*}

\section{Further experiments and results}

\subsection{Comparison of objectives}\label{app:compare-objectives}

In Section~\ref{sub:cbow}, we describe a more general training objective than
the classical CBOW objective from \cite{mikolov2013efficient}. The original
objective always sets the center word from the window of tokens $(w_{t - c},
\ldots, w_{t + c})$ as target word, $w_O = w_t$. In preliminary experiments,
this did not yield satisfactory results.
We believe that this objective is too simple for learning sentence embeddings
that capture semantic information. Therefore, we experimented a variant where
the target word is sampled randomly from a uniform distribution, $w_O := \mathcal{U} (\{w_{t-c},
\dotsc, w_{t+c} \})$.

To test the effectiveness of this modified objective, we evaluate it with the
same experimental setup as described in Section~\ref{sec:experiments}.
Table~\ref{app:tab:obj-probing} lists the results on the linguistic probing
tasks. CMOW-C and CBOW-C refer to the models where the center word is used as
the target. CMOW-R and CBOW-R refer to the models where the target word is
sampled randomly. While CMOW-R and CMOW-C perform comparably on most probing
tasks, CMOW-C yields 5 points lower scores on WordContent and BigramShift.
Consequently, CMOW-R also outperforms CMOW-C on 10 out of 11 supervised
downstream tasks and on all unsupervised downstream tasks, as shown in
Tables~\ref{app:tab:obj-downstreamS} and \ref{app:tab:obj-downstreamU},
respectively. On average over all downstream tasks, the relative improvement is
20.8\%. For CBOW, the scores on downstream tasks increase on some
tasks and decrease on others. The differences are miniscule. On
average over all 16 downstream tasks, CBOW-R scores 0.1\% lower than CBOW-C.
\begin{table*}[h]
\caption{Scores for different training objectives on the linguistic probing
tasks.}
\label{app:tab:obj-probing}
\resizebox{\textwidth}{!}{

\begin{tabular}{lrrrrrrrrrr}
\toprule
                                            Method &  Depth &  BShift & 
                                            SubjNum &  Tense & 
                                            CoordInv &  Length & 
                                            ObjNum &  TopConst &  SOMO &  WC \\
\midrule
CMOW-C &   \absbest{36.2} &         66.0 &        81.1 &   78.7 &                  
61.7 &    \absbest{83.9} &       79.1 &             73.6 &       50.4 &        
66.8
\\
 CMOW-R &   35.1 &         \absbest{70.8} &        \absbest{82.0} &  
 \absbest{80.2} & \absbest{61.8} & 82.8 &       \absbest{79.7} &            
 \absbest{74.2} & \absbest{50.7} & \absbest{72.9} \\
 \midrule
 \midrule
 CBOW-C &   \absbest{34.3} &         \absbest{50.5} &        \absbest{79.8} &  
 \absbest{79.9} & 53.0 & \absbest{75.9} &       \absbest{79.8} &            
 \absbest{72.9} & 48.6 & 89.0 \\
 CBOW-R &   33.0 &         49.6 &        79.3 &   78.4
& \absbest{53.6} & 74.5 &       78.6 &             72.0 &       \absbest{49.6} &        
\absbest{89.5} \\
\bottomrule
\end{tabular}

}
\end{table*}

\begin{table*}[h]
\caption{Scores for different training objectives on the supervised downstream
tasks.}
\label{app:tab:obj-downstreamS}
\resizebox{\textwidth}{!}{
\begin{tabular}{lrrrrrrrrrrr}
\toprule
                                            Method &  SUBJ &    CR &    MR & 
                                            MPQA &  MRPC &  TREC & 
                                            SICK-E &  SST2 &  SST5 &  STS-B & 
                                            SICK-R \\
\midrule
CMOW-C &  85.9 &  72.1 &  69.4 &  87.0 &  \absbest{71.9} &  85.4 &           
74.2 &  73.8 &  37.6 &          54.6 &             71.3 \\ 
 CMOW-R &   \absbest{87.5} &  \absbest{73.4} &  \absbest{70.6} &  \absbest{87.3}
 & 69.6 & \absbest{88.0} & \absbest{77.2} & \absbest{74.7} &  \absbest{37.9} &         
 \absbest{56.5} & \absbest{76.2} \\
 \midrule
 \midrule
 CBOW-C &  \absbest{90.0} &  \absbest{79.3} &  \absbest{74.6} &  \absbest{87.5}
 & \absbest{72.9} & 85.0 & \absbest{80.0} & 78.4 &  41.0 &         
 60.5 & \absbest{79.2}
 \\
 CBOW-R &   \absbest{90.0} &  79.2 &  74.0 &  87.1 &  71.6 &  \absbest{85.6} &           
 78.9 & \absbest{78.5} &  \absbest{42.1} &          \absbest{61.0} &            
 78.1 \\
\bottomrule
\end{tabular}}
\end{table*}

\begin{table*}[h]
\center
\caption{Scores for different training objectives on the unsupervised downstream
tasks.}
\label{app:tab:obj-downstreamU}
\begin{tabular}{lrrrrr}
\toprule
                                            Method &  STS12 &  STS13 &  STS14 &  STS15 &  STS16 \\
\midrule
CMOW-C &   27.6 &   14.6 &   22.1 &   33.2 &   41.6 \\
 CMOW-R &   \absbest{39.2} &   \absbest{31.9} &   \absbest{38.7} &  
 \absbest{49.7} & \absbest{52.2}
 \\
 \midrule
 \midrule
 CBOW-C &   \absbest{43.5} &   49.2 &   \absbest{57.9} &   \absbest{63.7} &  
 \absbest{61.6} \\
 CBOW-R &   \absbest{43.5} &   \absbest{50.0} &   57.7 &  
 63.2 & 61.0 \\
\bottomrule
\end{tabular}
\end{table*}

\subsection{Initialization strategy}\label{app:sec:init}

In Section~\ref{sec:initialization}, we present a novel random initialization
strategy. We argue why it is more adequate for training
CMSMs than classic strategies that initialize all parameters with random values
close to zero, and use it in our experiments to train CMOW.

To verify the effectiveness of our initialization strategy empirically, we
evaluate it with the same experimental setup as described in
Section~\ref{sec:experiments}. The only difference is the initialization
strategy, where we include Glorot initialization~\citep{glorot2010understanding}
and the standard initialization from $\mathcal{N} ( 0, 0.1 )$.
Table~\ref{app:tab:init-probing} shows the results on the probing tasks.
While Glorot achieves slightly better results on BShift and TopConst, CMOW's
ability to memorize word content is improved by a wide margin by our
initialization strategy. This again affects the downstream performance as shown
in Table~\ref{app:tab:init-downstreamS} and \ref{app:tab:init-downstreamU},
respectively:
7 out of 11 supervised downstream tasks and 4 out of 5 unsupervised downstream tasks
improve. On average, the relative improvement of our strategy compared to
Glorot initialization is 2.8\%.
\begin{table*}[h]
\caption{Scores for initialization strategies on probing tasks.}
\label{app:tab:init-probing}
\resizebox{\textwidth}{!}{

\begin{tabular}{lrrrrrrrrrr}
\toprule
                                            Initialization &  Depth &  BShift & 
                                            SubjNum &  Tense & 
                                            CoordInv &  Length & 
                                            ObjNum &  TopConst &  SOMO &  WC \\
\midrule
 $\mathcal{N} ( 0, 0.1 )$ &   29.7 &         71.5 &        82.0 &   78.5 &                   60.1
 & 80.5 &       76.3 &             74.7 &       \absbest{51.3} &         52.5 \\
 Glorot &   31.3 &         \absbest{72.3} &        81.8 &   78.7 &                  
 59.4 & 81.3 &       76.6 &             \absbest{74.6} &       50.4 &        
 57.0
 \\
 Our paper &   \absbest{35.1} &         70.8 &        \absbest{82.0} &  
 \absbest{80.2} & \absbest{61.8} & \absbest{82.8} &       \absbest{79.7} &            
 74.2 & 50.7 & \absbest{72.9} \\
\bottomrule
\end{tabular}

}
\end{table*}

\begin{table*}[h]
\caption{Scores for initialization strategies on supervised downstream tasks.}
\label{app:tab:init-downstreamS}
\resizebox{\textwidth}{!}{
\begin{tabular}{lrrrrrrrrrrr}
\toprule
Initialization &  SUBJ &    CR &    MR & 
                                            MPQA &  MRPC &  TREC & 
                                            SICK-E &  SST2 &  SST5 &  STS-B & 
                                            SICK-R \\
\midrule
$\mathcal{N} ( 0, 0.1 )$ &    85.6 &  71.5 &  68.4 &  86.2 &  \absbest{71.6} & 
86.4 & 73.7 & 72.3 &  \absbest{38.2} &          53.7 &             72.7 \\
Glorot &  86.2 &  \absbest{74.4} &  69.5 &  86.5 &  71.4 &  \absbest{88.4} &           
 75.4 & 73.2 &  \absbest{38.2} &          54.1 &             73.6 \\
Our paper &   \absbest{87.5} &  73.4 &  \absbest{70.6} &  \absbest{87.3} &  69.6 & 
88.0 & \absbest{77.2} & \absbest{74.7} &  37.9 &          \absbest{56.5} &            
\absbest{76.2}
\\
\bottomrule
\end{tabular}}
\end{table*}

\begin{table*}[h]
\center
\caption{Scores for initialization strategies on unsupervised downstream
tasks.}
\label{app:tab:init-downstreamU}
\begin{tabular}{lrrrrr}
\toprule
                                            Initialization &  STS12 &  STS13 & 
                                            STS14 & STS15 &  STS16 \\
\midrule
$\mathcal{N} ( 0, 0.1 )$ &   37.7 &   26.5 &   33.3 &   44.7 &   50.3 \\
Glorot &   \absbest{39.6} &   27.2 &   35.2 &   46.5 &   51.6 \\
Our paper &   39.2 &   \absbest{31.9} &   \absbest{38.7} &   \absbest{49.7} &  
\absbest{52.2} \\
\bottomrule
\end{tabular}
\end{table*}

\end{appendices}

\end{document}